# The Use of Multimodal Large Language Models to Detect Objects from Thermal Images: Transportation Applications

Huthaifa I. Ashqar, Taqwa I. Alhadidi, Mohammed Elhenawy, and Nour O. Khanfar

*Abstract*—The integration of thermal imaging data with Multimodal Large Language Models (MLLMs) constitutes an exciting opportunity for improving the safety and functionality of autonomous driving systems and many Intelligent Transportation Systems (ITS) applications. This study investigates whether MLLMs can understand complex images from RGB and thermal cameras and detect objects directly. Our goals were to 1) assess the ability of the MLLM to learn from information from various sets, 2) detect objects and identify elements in thermal cameras, 3) determine whether two independent modality images show the same scene, and 4) learn all objects using different modalities. The findings showed that both GPT-4 and Gemini were effective in detecting and classifying objects in thermal images. Similarly, the Mean Absolute Percentage Error (MAPE) for pedestrian classification was 70.39% and 81.48%, respectively. Moreover, the MAPE for bike, car, and motorcycle detection were 78.4%, 55.81%, and 96.15%, respectively. Gemini produced MAPE of 66.53%, 59.35% and 78.18% respectively. This finding further demonstrates that MLLM can identify thermal images and can be employed in advanced imaging automation technologies for ITS applications.

**Keywords:** MLLM, Thermal Images, RGB, GPT4, Gimini.

## I. Introduction

Thermal imaging is a critical component of the development of automatic driving systems. This can be seen in an example of pedestrian detection [1]. Deep learning methods, such as Convolutional Neural Networks (CNNs), have proven effective in automating object detection in thermal images [2], [3], [4]. These methods are important because they enable self-driving cars to detect and react to various objects and barriers in their environment. This has direct implications for the safety and economy of the transportation systems. Vehicles equipped with various sensor technologies, such as LiDAR, Cameras, and thermal images, can fuse their data into a unified sensor framework [5]. Thermal imaging and other sensing technologies combined with machine learning techniques facilitate object identification and lane-keeping enhancements, thus contributing to a safer and more comfortable riding environment for autonomous vehicles [6]. In the context of automated driving, thermal images may indicate the development of advanced driver assistance systems (ADAS) [7].

Multimodal large-language models (MLLMs) require multimodal capabilities to improve the effective function and safety of computer-driven systems. LLMs multimodal design using images, videos, and text from various data types has the potential to support collaborative learning scenarios [8]. They have also been applied to a variety of educational and clinical technologies with diverse applications [9], [10]. Moreover, the integration of language models with image understanding in multimodal LLMs has been extensively studied and is an indication of future trends [11]. Multimodal LLMs have shown great promise in the field of computerized driving. Vision-language pre-training (VLP) is a method for improving the performance of downstream vision and language tasks by pretraining models on a large number of image-text pairs, which is critical for tasks such as image captioning and visual question answering [12], [13]. Moreover, as large-scale vision-and-language pre-training has become a mainstream trend, these models have achieved state-of-the-art performance on several vision-and-language tasks, reflecting their potential for enhancing the capabilities of automated driving systems [14]. However, the adoption of automatic driving technology also depends on public opinion and attitudes towards such systems [15], [16], and the introduction of highly automated driving systems must consider issues such as complexity, familiarity, and human factors [17], [18], [19].

To this end, multimodal large-language models, including Gemini and GPT-4, are integrated in thermal image recognition under different scenarios. In this research paper, we articulate three primary research questions to explore the capabilities of the Multimodal Learning Language Model (MLLM):

1. Does MLLM possess a generalized understanding of images regardless of the camera type, such as RGB or thermal imaging? Can the MLLM effectively detect and identify various objects in thermal images for ITS applications?

2. Is MLLM capable of discerning whether two distinct images, RGB and thermal, are captured within the same scene?

3. Does MLLM comprehend the two image modalities and accurately detect objects in a scene based on both modalities for ITS applications?

The remainder of this paper is organized as follows: the next section summarizes some of the key papers on state-of-art, our proposed methodology is presented in Section 3, Section 4 presents the study results, and we discuss them in

H. I. Ashqar is with Arab American University Jenin, Palestine, and Columbia University, NY, USA (E-mail: huthaifa.ashqar@aaup.edu).

T. I. Alhadidi is with Civil Engineering Department, Al-Ahliyya Amman University, Amman, Jordan (E-mail: t.alhadidi@ammanu.edu.jo).

M. Elhenawy is with CARRS-Q, Queensland University of Technology Brisbane, Australia, (E-mail: mohammed.elhenawy@qut.edu.au).

N. O. Khanfar is with Arab American University, Jenin, Palestine, (E-mail: n.khanfar1@student.aaup.edu).

Section 5. Finally, we conclude our paper in Section 6 with a conclusion and directions for future research.

## II. LITERATURE REVIEW

The integration of Multimodal Large Language Models (MLLM) into thermal images is an emerging research direction with potential applications. Some multimodal models, such as CLIP, have demonstrated great potential in connecting imaging and natural language and fitting into thermal images, demonstrating that such models could reach new levels of downstream performance [20]. The multimodal guide also applies such information in the upstream stream to increase the performance of the visual downlink mission to enable the second stream to consider the image information contained in the acquisition of a multimodal machine learning model language that overcomes language variation depending on the classification of the object and the update of the online content [21]. The potential of large-scale dual-stream vision language pre-training, such as CLIP and ALIGN, has proven to be useful for the overall performance and downstream goals of various multimodal alignment levels, including image text recovery and imaging [22]. Multimodal transformer networks have shown the best performance with excellent cross-functional modality, ideal for various vision and linguistic tasks such as image text recovery and image indication [22]. Additionally, to have contextually relevant crosses of speech with crossed-image labels "tokenization" has three ideas because it supports the extrapolation of various multimodal alignment scopes to language-only scope data [23]. This idea would be suitable for incorporating thermal images into a large speech model integration, as it would facilitate a direct connection between visual and noun semantics. Finally, a large image dataset, such as GEM, is a viable option for researchers to align imaging artifacts at high- and multilingual labeling points to assign a perfect image for the first amendment to a multimodal model fit application [24].

In image interpretation and remote sensing, feature cameras achieve greater color accuracy than RGB technology, using a multispectral filter wheel [25]. Moreover, in drones, the combination of RGB-IR conservation aids in recognizing objects generated by transportation [26]. Moreover, RGB-D images can be used for segmentation to show reflections in an elevator environment, as the authors proved difficult to transport [27]. In geographic information sciences and computer vision, human orientation is estimated from the RGB photographs of users [28], [29]. Encryption and image secret writing are often combined using RGB images to protect or distort the data [30], [31].

Ironically, while text-based language models such as GPT-3 [32], BERT [33], and RoBERTa [34] outperform humans in text production and encoding tasks, their comprehension and processing capabilities are almost non-existent, considering that understanding data encompasses a variety of other types. Multimodal LLMs solve this issue by working with various data types and by introducing opportunities to work with other types of data by transcending data models that only work with text. GPT-4 has very few multimodal LLMs. According to researchers, GPT-4 is a MLLM that demonstrates human-level performance on major league benchmarks for image and text input [35]. The ability to perceive and understand inputs from multiple sensory modalities is a crucial aspect of AI development. It is critical to being able to learn and navigate successfully in the physical world. Additionally, multimodal inputs can significantly improve the LLM performance in downstream applications by enabling new fields beyond language, such as multimodal robotics, document intelligence, and robot technology.

## III. METHODOLOGY

### A. Dataset

We used the Teledyne FLIR Free ADAS Thermal Dataset v2, which is a comprehensive collection of annotated thermal and visible spectrum frames intended for the development of object detection neural networks. This dataset aims to promote research on visible + thermal spectrum sensor fusion algorithms ("RGBT") to enhance the safety of autonomous vehicles. It comprises 26,442 fully annotated frames covering 15 different object classes. The data was captured using a thermal and visible camera pair mounted on a vehicle, with the thermal camera operating in T-linear mode. Thermal images were acquired with a Teledyne FLIR Tau 2 13 mm f/1.0 camera, while visible images were captured with a Teledyne FLIR BlackFly S BFS-U3-51S5C (IMX250) camera. Time-synced capture was facilitated by Teledyne FLIR's Guardian software, enabling frame rates of 30 frames per second in validation videos, which also include target IDs for tracking metrics computation. The dataset ensures diversity in training and validation by sampling frames from a wide range of footage, with some frames selected manually and others using a frame skip. Redundant footage, such as identical frames during red light stops, was excluded by the curation team. Fig. 1 shows and example of the images and corresponding annotations

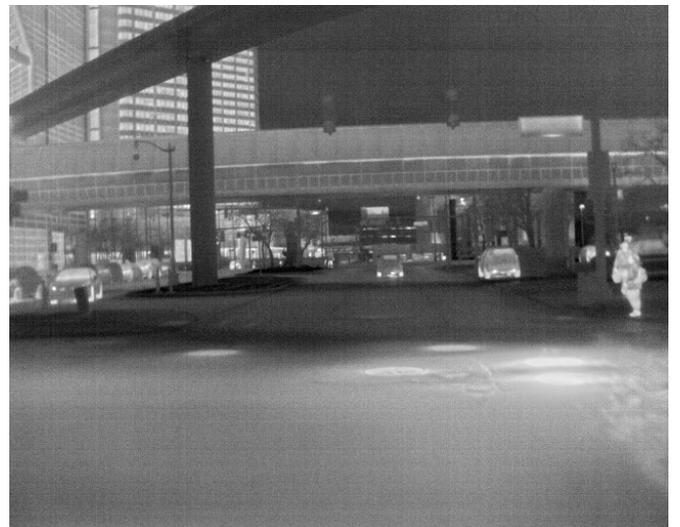

Annotation of the classes in the image: {1: 1, #Person; 2: 0, # Bike; 3: 4, # Car; 4: 0, # Motorcycle; 6: 0, # Bus; 7: 0, # Train; 8: 0, # Truck; 10: 0, # Traffic light; 11: 0, # Fire hydrant; 12: 0, # Street sign; 17: 0, # Dog; 37: 0, # Skateboard; 73: 0, # Stroller; 77: 0, # Scooter; 79: 0  # Other vehicle}

Figure 1.  Example of the images and corresponding annotations.

## B. Proposed Framework

The research methodology is illustrated in Fig. 2. To address the first question, we designed a study utilizing zero-shot in-context learning. MLLM was used to identify objects within a thermal image and to enumerate the occurrence of each object. This experiment was conducted using a selected subset of thermal training images.

To answer the second question, we experimented with thermal and RGB test images that have a one-to-one correspondence and have been captured in identical scenes. We employed the chain-of-thought technique in our prompt design, instructing the model to describe the RGB image followed by a thermal image. Subsequently, the model was used to estimate the likelihood of both the images originating from the same scene. This experiment was conducted in two iterations: the first with image pairs from the same scene and the second with image pairs from disparate scenes.

To answer the third question, we leveraged the test data with established mapping information and crafted a prompt directing the model to recognize objects and count their presence using dual-image modalities. Lastly, to quantify the improvement in object detection, we repeated the detection experiment twice, each time utilizing a single image modality for comparison.

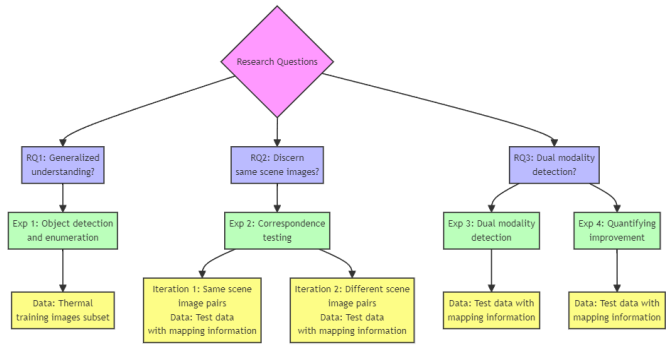

Figure 2. Flowchart of the proposed methodology.

## IV. RESULTS

### A. Capability Across Different Images

The main goal of the first experiment was to investigate the potential for generalizing MLLM knowledge across thermal and RGB imaging modalities. Zero-shot in-context learning methods indicate that the models can handle the process and analyze modalities. However, the accuracy of item detection varied among photos. In particular, thermal images present a special challenge because they depend on thermal traces rather than visible light and contain less visual information. However, MLLM was able to accomplish moderate object recognition, including cars and people, demonstrating a solid starting point for future model development.

### B. Categorization and Recognition of Objects

The model also performed significantly well in recognizing and identifying objects. The confusion matrices for Gemini and GPT4 are shown in Fig. 3 and Fig. 4, respectively. The True Positive Rate for vehicle detection was 0.86, while that for motorcycle detection was 0.08. By contrast, RGB images contain more obvious visual hints, making them suitable for spotting smaller items. However, the modalities did a poor job of recognizing objects in scenarios with many objects or many backgrounds.

### C. Analyzing Scene Consistency

The second series of experiments revealed the following additional issues. The model performed better when using two thermal images of the same scene and photos from two different image sensors, one with RGB and the other with IR. Using a chain-of-thought process, the models generated a comparison of photos from the same scenes and performed identical and diversified scenes. Accurate identification proved to be moderately adequate, with a recall of 0.91 and a precision of 0.79 for similar scenes. However, its performance deteriorated between scenes, falling to a recall of 0.57 and a precision of 0.79. This variance illustrates the difficulty of the models in understanding their settings and distinguishing between identical scene structures using unique modality views. A comparison between the two models is presented in TABLE I.

TABLE I. COMPARIOSN BETWEEN GEMINI AND GPT4.

| Evaluation Metrics | Gemini 1.0 Pro Vision | GPT4 Vision Preview |
|---|---|---|
| Precision (same scene) | 0.79 | 0.50 |
| Recall (same scene) | 0.91 | 0.64 |
| F1 Score (same scene) | 0.85 | 0.56 |
| Precision (different scene) | 0.79 | 0.50 |
| Recall (different scene) | 0.57 | 0.36 |
| F1 Score (different scene) | 0.66 | 0.42 |

### D. Object Detection Using Dual Modality

The final experiment demonstrated the power of combining the thermal and RGB data. The models were able to integrate both modalities to better visualize the image, resulting in greater object detection and even more precise detection under more complicated circumstances. In various traffic scenarios, this approach enhances the probability and operational performance of autonomous systems, thereby demonstrating that multisensory data integration is superior.

### E. Assessment Approach and Key Performance Indicators

All object categories were utilized to conduct an extensive model assessment using MAPE and MAE. According to the findings, the capacity to identify larger items such as cars and buses was far superior, demonstrating a substantial gap between categories. Because thermal and RGB sensors identify distinct characteristics, the performance disparity varies depending on the item category and situation. These results indicate that the models face challenges in accurately determining whether two images are captured in the same scene, primarily because of scene similarities, suboptimal lighting conditions, glare, or adverse weather conditions.

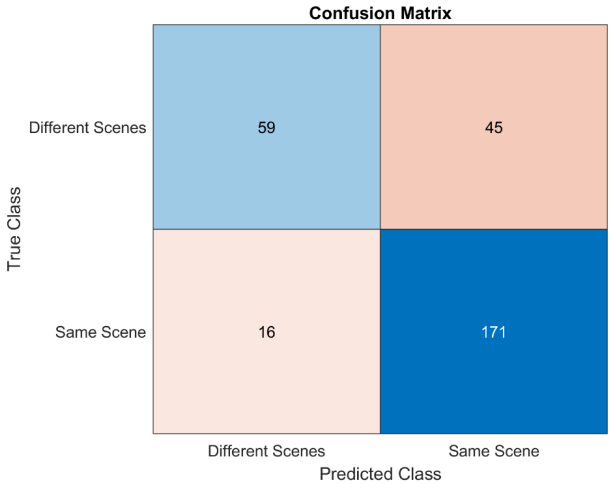

Figure 3. Gemini confusion matrix.

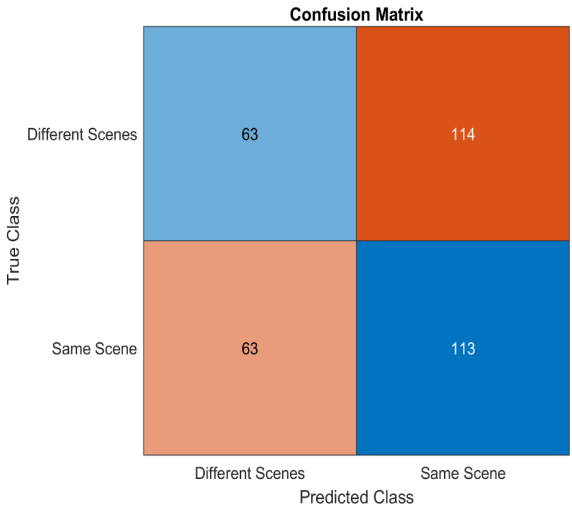

Figure 4. GPT4 confusion matrix.

*F. Evaluation Methodology*

Our evaluation methodology encompasses two primary aspects: accuracy of category detection and precision of enumeration within these categories. For this analysis, the successful identification of at least one object within a given category is deemed sufficient for category detection without considering the total number of objects detected.

To assess enumeration accuracy, we employed two statistical measures: the Mean Absolute Error (MAE) and Mean Absolute Percentage Error (MAPE). These metrics were applied to each category of the annotations. In scenarios where a category is inferred by the MLLM but does not exist in the annotations, the instance or image is omitted from the MAE and MAPE calculations to avoid skewing the results. These instances were considered false positives.

Given the possibility of false positives, we adopted a binary approach to quantifying object presence within categories: a category was assigned a value of one if it contained at least one object and zero otherwise. This binary system facilitates direct comparison between the inferred data and annotations, enabling us to accurately calculate the True Positive Rate (TPR) and False Positive Rate (FPR) for each category. This methodological framework ensures a balanced evaluation of the detection efficacy and enumeration accuracy.

TABLE II provides a performance breakdown of two MLLMs; GPT-4 Vision Preview and Gemini 1.0 Pro Vision, when analyzing thermal images. These results were not intended to serve as a direct comparison between the two models; rather, the goal was to conduct a comprehensive analysis covering a wide range of images. This was to ensure that the models performed similarly and to demonstrate the potential of utilizing MLLM models with thermal imaging data. Given that these models are not freely available, a methodical selection process involving random sampling of images is required for zero-shot in-context learning for each model, thereby optimizing the utility derived from their application. This approach is used to maximize the value received from their use.

From TABLE II, we can observe consistency in the performance across both models. This suggests that regardless of the model used, the ability to interpret the thermal data is solid. Interestingly, there appears to be a trend where the larger the object, the better the models' enumeration and detection capabilities, with the car category showing high True Positive Rates (TPR) for both models. Moreover, both models achieved zero False Positive Rates (FPR) in detecting motorcycles, and the Mean Absolute Error (MAE) and Mean Absolute Percentage Error (MAPE) scores across categories indicated promising precision and recall, especially for larger objects such as cars.

TABLE II. PERFORMANCE BREAKDOWN OF GPT4 AND GEMINI ANALYZING THERMAL IMAGES.

| MLLM Model | GPT4 Vision Preview | | | | Gemini 1.0 Pro Vision | | | |
|---|---|---|---|---|---|---|---|---|
| *Evaluation Metrics per Category* | *TPR* | *FPR* | *MAE* | *MAPE* | *TPR* | *FPR* | *MAE* | *MAPE* |
| Category ID 1: Person | 0.57 | 0 | 4.76 | 70.39 | 0.39 | 0 | 4.47 | 81.48 |
| Category ID 2: Bike | 0.31 | 0.01 | 1.89 | 78.40 | 0.41 | 0.02 | 1.37 | 66.53 |
| Category ID 3: Car (includes pickup trucks and vans) | 0.86 | 0.09 | 4.35 | 55.81 | 0.90 | 0.08 | 5.04 | 59.35 |
| Category ID 4: Motorcycle | 0.08 | 0 | 1.38 | 96.15 | 0.24 | 0.01 | 1.06 | 78.18 |

## V. Discussion and Conclusion

The accuracy of object classification in images from multiple modalities was assessed by measuring the MAPE values of the GPT4 and Gemini. GPT4 and Gemini demonstrated variable proficiency levels across several item categories, such as pedestrians, bicycles, automobiles, and motorcycles. Gemini had a higher level of precision in identifying pedestrians, with a MAPE of 81.48%, while GPT4 attained a MAPE of 70.39%. Therefore, Gemini is better suited to autonomous driving technologies related to pedestrians. The GPT-4 exhibited varying degrees of precision across distinct object categories. Significantly, motorcycles had a high MAPE of 96.15%, indicating that there is potential for improvement in identifying smaller or less easily recognizable objects in thermal pictures.

The ability of both models to generalize across RGB and thermal images is crucial to the robustness of autonomous systems under various lighting and weather conditions. The results indicate that, while generalization is possible, the differences in error rates across various object types and models imply that more calibration and training may be necessary. The safety of autonomous vehicles is highly dependent on accurate detection and identification of objects, particularly in complex environments. The existence of a somewhat high MAPE in specific categories implies that there might be instances where the models fail to accurately detect or misclassify objects, leading to hazardous driving decisions.

The results suggest the need for ongoing enhancements to enhance the accuracy of the model, especially in diverse environmental conditions that autonomous vehicles may encounter. Potential advancements and future uses: The findings endorse the ongoing use and enhancement of MLLMs for image-based processing in autonomous driving. The precision and reliability of the model training can be enhanced by integrating a broader array of images and environmental conditions. Moreover, these findings highlight the potential of employing these technologies in several domains that require robust and adaptable image recognition and processing skills, such as security and surveillance, environmental monitoring, and related fields.